\pdfoutput=1

\documentclass[11pt]{article}

\usepackage[]{ACL2023}

\usepackage{times}
\usepackage{latexsym}
\usepackage{graphicx}

\usepackage[T1]{fontenc}

\usepackage[utf8]{inputenc}

\usepackage{microtype}

\usepackage{inconsolata}
\usepackage[acronym]{glossaries}

\newacronym{ml}{ML}{Machine Learning}
\newacronym{llm}{LLM}{Large Language Model}
\newacronym{sota}{SotA}{state-of-the-art}
\newacronym{nlp}{NLP}{natural language processing}
\newacronym{ai}{AI}{Artificial Intelligence}
\newacronym{dl}{DL}{deep learning}
\newacronym{rl}{RL}{Reinforcement Learning}
\newacronym{gai}{GenAI}{Generative AI}
\newacronym{cot}{CoT}{chain-of-thought}
\newacronym{asi}{ASI}{artificial superintelligence}
\newacronym{agi}{AGI}{artificial general intelligence}
\newacronym{laws}{LAWS}{lethal autonomous weapon systems}
\newacronym{av}{AV}{autonomous vehicle}
\newacronym{nasa}{NASA}{National Aeronautics and Space Administration}
\newacronym{nn}{NN}{Neural Network}
\newacronym{ann}{ANN}{Artificial Neural Network}
\newacronym{bdi}{BDI}{Belief-Desire-Intention}
\newacronym{oecd}{OECD}{Organisation for Economic Co-operation and Development}
\newacronym{rlhf}{RLHF}{Reinforcement Learning from Human Feedback}

\title{AI Must not be Fully Autonomous}

\author{\\Tosin Adewumi*, Lama Alkhaled, Florent Imbert, Hui Han, Nudrat Habib, and Karl Löwenmark
\\
Machine Learning Group, EISLAB,
\\ Luleå University of Technology, Sweden \\
\footnotesize{*corresponding author}, 
\large
{\fontfamily{pcr}\selectfont
firstname.lastname@ltu.se}
}

\begin{document}
\maketitle
\begin{abstract}
Autonomous \acrfull{ai} has many benefits.
It also has many risks.
In this work, we identify the 3 levels of autonomous \acrshort{ai}.
We are of the position that \textit{\acrshort{ai} must not be fully autonomous} because of the many risks, especially as \acrfull{asi} is speculated to be just decades away.
Fully autonomous \acrshort{ai}, which can develop its own objectives, is at level 3 and without responsible human oversight.
However, responsible human oversight is crucial for mitigating the risks.
To ague for our position, we discuss theories of autonomy, \acrshort{ai} and agents.
Then, we offer 12 distinct arguments and 6 counterarguments with rebuttals to the counterarguments.
We also present 15 pieces of recent evidence of \acrshort{ai} misaligned values and other risks in the appendix.

\end{abstract}

\section{Introduction}

While \acrfull{ai} has many benefits \cite{sharma2024benefits,mon2025embodied,löwenmark2025agentbasedconditionmonitoringassistance,adewumi2025findings}, it also has its challenges \cite{pettersson2024generative,adewumi2024limitations,chakraborty2025heal}.
The primary focus of this position paper are the risks of misaligned values in \acrshort{ai} systems that learn, though we present existential threat and other risks as well.
Misaligned values may have become one of the most pressing challenges with \acrshort{ai} besides previous well-known challenges like social bias.
Some misaligned values include (1) deception \cite{meinke2024frontier,ren2025mask}, (2) alignment faking (i.e. selective compliance to avoid modification) \cite{greenblatt2024alignment}, (3) reward hacking (i.e. achieving high rewards through unintended behaviors in \acrfull{rl}) \cite{baker2025monitoring},  and (4) blackmail \cite{anthropic2025}.
Given these risks with frontier models and other \acrshort{ai} systems that learn, we are of the position that \textit{\acrshort{ai} must not be fully autonomous}.
Notice that we are not against autonomous \acrshort{ai} but fully autonomous \acrshort{ai}, thereby advocating for \textit{responsible human oversight}.


For context, we begin by defining some of the key terms that are relevant for this paper.
\textbf{\acrshort{ai} is broadly defined as the simulation of human intelligence in machines} \cite{dong2020research,lavery1986artificial,mccorduck1977history}.
According to the influential textbook by \citet{russell2016artificial}, \acrshort{ai} is the study of agents that receive percepts from the environment and perform action.
While their \acrshort{ai} definition does not seem to differentiate \acrshort{ai} from agent, \citet{wooldridge1994agent} define an agent as an autonomous and logical (or rational) entity.
This definition of an agent may be further clarified as an \acrshort{ai} entity that perceives its environment through data inputs, makes decisions based on rules or logic, and takes action through outputs to achieve specific goals \cite{castelfranchi1998modelling,wooldridge1999intelligent,vanneste2024artificial}. 
Apparently, the English word \textit{agent} was derived from the Latin equivalent and refers to someone or something that produces an effect \cite{minkova2009english}.
Discussions about the issue of full \acrshort{ai} autonomy is timely and important because of the growing cases of misaligned values while \acrshort{ai} is enjoying growing adoption at the same time, even as agents.

The approach in this paper starts by discussing the background and current paradigms about autonomy, \acrshort{ai}, and agents by identifying the theories around them (Section \ref{background}).
To make this work accessible to a much wider audience, we refrained from using mathematical equations in the Section.
We then present our 12 core arguments for the perspective we have taken (Section \ref{argument}).
These include \textit{existential threat}, \textit{inductive \acrshort{ai} inherits human attributes}, \textit{\acrshort{ai} bias and systemic prejudice}, \textit{\acrshort{ai} side-stepping human control}, and \textit{rise in the number of new \acrshort{ai} risks}, among others.
These are then followed by counterarguments in the literature to our perspective, for which we also provide rebuttals (Section \ref{counterargs}).
We also discuss the implications and future directions for the perspective we have provided (Section \ref{implications}).
Finally, we give concluding remarks with a call to action (Section \ref{conclusion}).

\paragraph{Contributions}
Previous work that align with our perspective or discuss the issue are somewhat limited.
Our work clearly articulates a position supported by growing evidence of \acrshort{ai} misaligned values.
Hence, the contributions of this paper can be highlighted as the following:


\begin{enumerate}
    \item The work gathers and presents 15 pieces of evidence of recent \acrshort{ai} misaligned values and other risks that cut across different fields.
    
    \item The work provides compelling arguments using relevant theories, counterarguments and rebuttals for our position.
    
\end{enumerate}



\section{Related work}
\label{related}

From the analysis by \citet{mitchell2025fully}, it was observed that  risks to people increase with the autonomy
of \acrshort{ai} systems while the sense of agency decreases for humans, according to \citet{fernandez2021trustworthy}.
\citet{kovac2022autonomous} recommended regulatory intervention over potential uncontemplated \acrshort{ai} risks.
\citet{o2023banning}, like other advocates, supported a ban on \acrshort{ai}-based \acrfull{laws}, especially since it's unknown how to make them reliably safe.
As a result, retaining agency over \acrshort{ai} through clear insight of the decision-making process was suggested \cite{o2023banning,sharkey2019autonomous}.
\citet{arkin2016ethics} was of the view that autonomous \acrshort{ai} is advancing faster than humans are able to understand all its implications or create adequate policies and legislation for its governance.
\citet{ebert2019validation} identified the challenges of validating autonomous systems and discussed both traditional and intelligent approaches for validation.

Due to the many advantages that \acrshort{ai} offers, \citet{walsh20182062} appeared to predict humans will be replaced by \acrshort{ai}.
It has been suggested that we may be on the verge of the \textit{singularity}, i.e. the hypothetical scenario of uncontrollable technological growth driven by \acrfull{asi} because, according to most \acrshort{ai} experts (out of 300), the chances that \acrshort{ai} will think like humans (or achieve \acrfull{agi}), is 50\% in 2062 and 90\% in 2220 \cite{walsh20182062}.
\citet{liu2023ai} actually advocated for \acrshort{ai} to be fully autonomous.
Their proposal was based on existing \acrfull{ml} paradigms and is thus susceptible to the pitfalls that existing \acrshort{ai} systems have.
For example, they identified, at least, two possible risks in their proposal: (1) achieving performance goals when making incorrect decisions and (2) safety in the adaptation process.
\citet{riesen2022moral} made a moral case for the use of fully autonomous weapons that can "\textit{reliably adhere to the laws of war}" but failed to state whose laws of war and how to validate this reliability.

\section{Background and Current Paradigms}
\label{background}

What constitutes \acrshort{ai} is a subject of much debate \cite{kelly2023factors}.
Perhaps, more so is the term agent.
It is argued that one of the reasons for the lack of consensus is that different elements associated with agency (e.g. learning or environment) have different levels of importance in different domains \cite{russell2016artificial}.
It can, therefore, help to go back to the earliest use of the term \acrshort{ai}.
\acrshort{ai}, as a term and field of research, was coined by a team of scientists, including John McCarthy, Marvin Minsky, Nathaniel Rochester and Claude Shannon, in 1955, when they proposed that "\textit{...any other feature of intelligence can in principle be so precisely described that a machine can be made to simulate it}" \cite{mccarthy2006proposal}.
Some researchers question the use of the term \acrshort{ai}, arguing that anthropomorphizing intelligence to machines is inappropriate \cite{totschnig2020fully,yudkowsky2001creating}.
There is nothing strange, however, about anthropomorphism (or personification) as it is a common metaphorical expression in many cultures because it is an innate tendency of human psychology \cite{hutson20137}.
\citet{walsh20182062} even went further to label \acrshort{ai} as \textit{Homo Digitalis}, in clear similarity to Homo Sapiens, which refers to humans.

\acrshort{ai} may be divided into 3 categories on the basis of task scope and performance \cite{kelly2023factors,bory2024strong,johnsen2025super}.

\begin{enumerate}
    \item Narrow (or weak) - This focuses on a specific task (e.g. image classification).
    
    \item General (i.e. \acrshort{agi}) - This combines multiple tasks with performance around average human level.
    
    \item Superintelligent (i.e. \acrshort{asi}) - This is conceived as capable of performing a broad array of tasks at levels exceeding human intelligence and with autonomous self-improvements.
\end{enumerate}

To better discuss \acrshort{ai} and agents using relevant theories (i.e. established explanations), we shall first discuss theories of autonomy, since this is an important attribute of agents \cite{wooldridge1994agent}.

\subsection{Theories of Autonomy}
\label{autonomy}

Autonomy is to self-govern. In philosophy, it is the ability to decide one's goal of action \cite{totschnig2020fully}, as alluded to from its Greek derivative \textit{autonomia} - "give oneself the law".
Hence, in \acrshort{ai}, autonomy involves the absence of direct, real-time human intervention \cite{hauer2020machine}.
In the context of space exploration, the U.S. agency, \acrfull{nasa}, defines autonomy as the ability to function independently to achieve goals \cite{zeigler1990high}.
A goal is the mental (or internal anticipatory) representation of a world state of action results  \cite{castelfranchi1998modelling}.
Below, we identify some key philosophical theories of autonomy.

\begin{enumerate}

    \item Procedural autonomy: The necessary and sufficient condition for this is to act on the desire one desires to have without attention to the values contained in it \cite{formosa2021robot}.
    Some are of the view Procedural autonomy is insufficient since it is "content-neutral".

    \item Substantive autonomy: This addresses the concerns of procedural autonomy by holding that autonomy is based on action from the right set of values or valid norms
    \cite{formosa2021robot}.
    Unlike Procedural autonomy, the content of one's values matter in this case.
    Critics of this theory argue that it is unclear which values are the right ones \cite{formosa2021robot}.

    \item Kantian autonomy:
    \citet{kant2020groundwork} argues that to be rational is to act autonomously.
    It is the capacity of an agent to act based on objective morality (seen as universal principle) instead of desires \cite{hooker2019truly}.
    According to \citet{formosa2021robot}, Kantian views fall under Substantive autonomy.

    \item Relational autonomy: This contextualizes autonomy within the social framework of interpersonal relations and mutual dependencies. \cite{christman2004relational}.
    It is a concept in feminist philosophy that captures a more social picture of agency.
    
\end{enumerate}

\paragraph{Levels of autonomy:}
Autonomy has to be understood as a relative term.
For humans, internal (e.g. genes) and external factors (e.g. environment) condition our goals and actions \cite{hooker2019truly,formosa2021robot}.
Autonomy, therefore, can be high or low and increase or decrease.
\citet{shrestha2021nature} differentiates automated \acrshort{ai} from autonomous \acrshort{ai} and the potential harm from each.
Autonomous \acrshort{ai} makes decisions by itself and learns from the environment unlike automated \acrshort{ai}, which is a product of design or coding.
\citet{fernandez2021trustworthy}, however, view autonomous as the highest level of automation (in a six-level model) in the domain of automated/\acrfull{av}, requiring no driver supervision.
\citet{zeigler1990high} identified 3 levels of autonomy in their model-based architecture for intelligent control (Table \ref{levels}).
This model with 3 levels is simple and captures the possibility of an agent to develop its own objectives unlike the 6-level model of \citet{fernandez2021trustworthy}.
Fully autonomous \acrshort{ai} is the \textbf{\acrshort{ai} at level 3 without responsible human oversight}.

\begin{table}[h!]
\small
\centering
\caption{Levels of autonomy \cite{zeigler1990high}}
\label{levels}
\begin{tabular}{p{0.1\linewidth} | p{0.7\linewidth}}
\hline
      \textbf{Level}   & \textbf{Description} \\
      \hline
      1   & Involves achievement of set objectives.\\
      2   & Involves the ability to adapt to changes in the environment.\\
      3   & Involves the ability of the system to develop its own objectives. This is the highest level.\\
 \hline
\end{tabular}
\end{table}

\subsection{Theories of \acrshort{ai}}
\label{aitheories}

It is sometimes argued that \acrshort{ai} has no widely accepted theory and, therefore, suffers from internal fragmentation \cite{wang2012theories}.
In spite of this, below, we identify some key theories of \acrshort{ai} by specifying a main theory (and a relevant theory under it).

\begin{enumerate}
    \item Cognitive science (Symbolic logic):
    Cognitive science is an interdisciplinary study of mind and intelligence \cite{thagard2005mind}.
    It proposes mental procedures operating on mental representations for producing thought and action.
    Symbolic logic (or formal deductive reasoning), which is "\textit{systematic common sense}" with the representation of symbols as arguments, forms an important part of cognitive science.
    One of the problems with symbolic systems is that they lack generalization capability.

    \item Connectionism (\acrfull{nn}):
    Connectionism is based on modeling the human brain and is useful for difficult problems in cognition.
    According to \citet{rumelhart1998architecture}, there are 7 main components for any connectionist system, like an \acrfull{ann}.
    These are (1) a processing unit, (2) an activation function, (3) an output function per unit, (4) connectivity among units, (5) an activation for combining the inputs, (6) a learning rule, and (7) an environment where the system operates.
    Connectionist systems are good for similarity-based generalization through inductive reasoning.
    
    \item Decision theory (Probability theory):
    Decision theory gives a rational framework for making choices among alternatives during uncertainty \cite{north2007tutorial}.
    Its foundations are based on utility theory (or value) and probability theory.
    Probability theory encodes the information about which outcome is likely to occur.
    For the edge cases, when we are certain about an alternative that will occur, the probability is 1 while it is 0 when it is certain not to occur.
    Many real-life situations, however,  are usually uncertain.
    The sum of probabilities of the mutually exclusive alternatives should always be 1.
    The avenue by which probability distributions are updated to include new information (based on the condition of past occurrences) is called Baye's rule \cite{north2007tutorial}.
    As with humans, \acrshort{ai} has to decide among alternatives in uncertain situations.
    
    \item Optimization theory (Evolutionary computation):
    Evolutionary computation is a major field of system optimization and draws inspiration from natural evolution \cite{floreano2008bio}.
    Its objective is to find solutions to hard optimization problems.
    It inherits the 4 main pillars of natural evolution: (1) population maintenance, (2) diversity creation, (3) selection mechanism, and (4) a process of genetic inheritance.
    A fitness function, which associates a score to each "phenotype" in the population, attempts to optimize the objective of the problem, as it undergoes the selection process.
    Examples of evolutionary computation algorithms include \textit{genetic algorithms}, \textit{island models}, and \textit{simulated annealing} \cite{floreano2008bio}.
    
    \item Control theory (\acrfull{rl}):
    Control theory is a theory of designing complex actions from well-specified models \cite{recht2019tour}.
    A central idea of control theory is the control loop, involving (1) the \textit{observation} from the environment, (2) \textit{inputs} applied to the system or agent, and (3) a mapping, called \textit{policy} or feedback law, from the observation to the input of the agent   \cite{meyn2022control}.
    \acrshort{rl} rests on the pillars of control theory and information theory.
    It involves the use of past data to optimize the future manipulation of a dynamical system.
    There are several variants (e.g. model-based and model-free, which includes policy search and approximate dynamic programming).
    \acrshort{rl} can be rewarding but its complicated feedback loops are hard to study in theory \cite{recht2019tour}.
    
\end{enumerate}

\subsection{Theories of Agent}
\label{agenttheories}

Many of the theories of \acrshort{ai} apply to agent. Hence, we briefly discuss a couple of agent-specific theories below.

\begin{enumerate}
    \item Theory of Mind: In the field of psychology, this is the representation of mental states (thoughts and goals) of others \cite{leslie2004core}.
    It frames the \acrfull{bdi} attribution.
    There are 2 prevailing models for \acrshort{bdi}: (1) selection by inhibition (for effective reasoning about belief) and (2) inhibition of return (involving 2 inhibitions applied serially).
    It has been questioned if the 3 attitudes of \acrshort{bdi} are sufficient, especially in multi-agent systems.

    \item Game theory: It is the modeling and analysis of interactive decision-making that has application in many fields, including Mathematics and Economics \cite{maschler2020game}.
    The decision of each agent affects the outcome for all the other agents in a multi-agent system.
    Decisions are taken within the general idea of a "game", involving (1) alternating moves, (2) limited knowledge, and (3)  an utility function, such that the higher the value, the more favorable  it is \cite{owen2013game}. 
    
\end{enumerate}

\paragraph{Classes of Agents:}
Given that autonomy is a relative term, it follows that \acrshort{ai} agents can be classified into categories.
We identify 5 categories based on the level of intelligence and interaction with the environment \cite{wooldridge1999intelligent,yudkowsky2001creating,russell2016artificial}.

\begin{enumerate}
    \item Simple reflex agent.
    This is the most basic and is only reactive.
    It follows predefined condition-action rules.
    A common example that is usually given is the thermostat.
    
    \item Model-based reflex agent. 
    This is also based on condition-action rules but has an internal model of the world to track the state of its environment.
    An example can be a simple robot.
    
    \item  Goal-based agent.
    This incorporates a goal-oriented approach to its actions by planning before acting with the ultimate aim of achieving its goal.
    An example is a robot designed with path planning.

    \item Utility-based agent.
    It relies on an utility function to assess and decide from a range of possible actions the one that gives the maximum benefit.
    An example can be an \acrshort{av} with the utility to maximize fuel economy when given a destination as a goal.

    \item Learning agent. 
    A learning agent improves its performance over time through experiences.
    Learning agents are designed to handle complexities beyond what is applicable in goal-based and utility-based agents.
    An example can be an \acrshort{av} with the learning elements.
    


\end{enumerate}

\paragraph{Agent's environment}
An agent may be situated in different environments and the complexity of the agent's decision-making process can be affected by the type of environment.
According to \citet{wooldridge1999intelligent}, some of these include:

\begin{enumerate}
    \item Accessible vs inaccessible - Where the agent can obtain complete and accurate information about the state of the environment, it is accessible. Different environments may be accessible to different levels. It is easier to build agents for more accessible environments than for complex or inaccessible ones.
    
    \item Deterministic vs non-deterministic - A single guaranteed effect is the outcome of any action for a deterministic environment. Uncertainty governs non-deterministic environments and this presents greater problems for agent designers.
    
    \item Static vs dynamic - An unchanging environment that only changes based on the action of an agent is considered static while a dynamic environment has other ongoing processes and is shaped by actions besides the agent's own.  
\end{enumerate}


\section{Core Arguments}
\label{argument}

The position we hold may appear too strong to some.
However, there are very strong reasons for this.
Beyond hypothetical conjectures, recent experiences and research \cite{jaech2024openai,meinke2024frontier,anthropic2025} have demonstrated strong support in favor of our position.
Furthermore, in the appendix (\ref{sec:appendix}), we present a list of examples of some of the experiences from different reliable sources, involving \acrshort{ai} misaligned values and other risks.
The examples on cases around deception are such that they may be regarded as deeper than the error of hallucination because the \acrshort{ai}, in such cases, insisted on its original answer when queried about it.
Below, we argue for reasons \acrshort{ai} must not be fully autonomous.
Some may suggest a time limit for this position after which the position can be dispensed off and \acrshort{ai} can be made fully autonomous.
However, since some of the arguments discussed below have no time limitation to their validity, we do not subscribe to the idea of a time limit.
The first argument is on possibly the most consequential risk for humanity - \textit{existential threat}.
Arguments \ref{inductive} to \ref{ethical} border on misaligned values and the remaining 4 arguments address other risks.

\subsection{Existential threat}
\label{existential}

The potential harm to humanity from \acrshort{ai}'s possibly misaligned goals is seen as an existential threat \cite{johnsen2025super,sparrow2024friendly}.
\citet{totschnig2020fully} argues that it is indeed possible for \acrshort{ai} to modify it's goal from the original that may have been given it.
As discussed in Section \ref{autonomy} with \citet{zeigler1990high}'s model of 3 levels of autonomous agents, this is the highest.
Real-life instances of agents modifying their goal have recently been observed, as pointed out by \citet{meinke2024frontier} with frontier models.
If an \acrshort{ai} system modifies its original human-assigned goal to another that is detrimental to humanity (or even both humanity and \acrshort{ai}), then humanity faces an existential threat.
This idea is even more disturbing when we consider that \acrshort{ai} is being considered in the military for \acrshort{laws} \cite{ilachinski2017artificial,o2023banning,asaro2020autonomous}.
\acrshort{laws} with level 2 autonomy without responsible human oversight is sufficient as existential threat.
It is for this reason \citet{sharkey2019autonomous} set out the parameters for "meaningful human control" of \acrshort{laws}, which includes retention of the power to suspend an attack during conflicts, and over 4,900 \acrshort{ai} and robotics researchers (besides 27,800 others) have signed an open letter calling for a ban on \acrshort{laws} that are beyond meaningful human control.\footnote{https://futureoflife.org/open-letter/open-letter-autonomous-weapons-ai-robotics/ - includes Stephen Hawking, Noam Chomsky, Geoffrey Hinton and more.}
\citet{heyns2017autonomous} extends this concept to scenarios outside of \acrshort{laws} by asserting that meaningful human control should be a guiding principle for \acrshort{ai} generally.
Our position of \textit{responsible human oversight} aligns with theirs.

\subsection{Inductive \acrshort{ai} inherits human attributes}
\label{inductive}

Machines were originally conceived to simulate human intelligence but it appears they can simulate more than just intelligence.
We here offer one argument collectively for "bad" or "unacceptable" human attributes that inductive \acrshort{ai} (identified in Section \ref{aitheories}) inherits, though we could make the case that each attribute can stand as an argument on their own and we do make the case for a few in the coming subsections.
What is "acceptable" or "unacceptable" can be debatable in certain circumstances or cultures, there are some universally accepted ones (e.g. as expressed through the universal declaration of human rights) \cite{donnelly1984cultural}.
Human attributes such as bias, prejudice or deception are usually considered unacceptable in many circumstances, if not all the time.
Such attributes may be embedded in the training data or algorithms of \acrshort{ai} systems.
For example, \citet{greenblatt2024alignment} showed that some models can exhibit "fake alignment" (a form of deception) by appearing helpful while secretly pursuing different goals.
This is particularly dangerous in autonomous \acrshort{ai} where hidden behaviors may only surface after deployment.
Given this collective risk, it is crucial to have responsible human oversight over autonomous \acrshort{ai}, especially as these bad attributes have not disappeared in humans generation after generation.

\subsection{\acrshort{ai} bias and systemic prejudice}
\label{bias}

An often underestimated risk of \acrshort{ai} systems is their capacity to reproduce and amplify social biases, particularly those relating to race, gender, and class \cite{alkhaled2023bipol}.
As many \acrshort{ai} models are trained using large-scale datasets derived from internet content, institutional records, or user behavior, they inherently reflect the inequalities embedded in these data sources.
There have been many studies on \acrshort{ai} bias.
For example, \citet{pmlr-v81-buolamwini18a} revealed that commercial facial recognition systems demonstrated severe accuracy disparities across race and gender, 
\citet{10.1007/978-3-031-23618-1_1} showed that gender stereotyping in facial expression recognition 
disproportionately associates assertive or negative emotion with male-presenting face and more submissive or positive emotion with female-presenting face, and  
\citet{Chen2023} demonstrated how algorithmic bias affects \acrshort{ai}-based recruitment systems, as
replicate discriminatory practices, particularly towards candidates from marginalized groups.
A safe way to completely eliminate such biases is unlikely as long as current paradigms, like connectionism and learning from data, remain or decisions are made using sensitive attributes (e.g. race or gender) \cite{ferrara2023should}.
Fully autonomous \acrshort{ai} operating without responsible human oversight 
risks perpetuating these injustices uncontrolled.
Therefore, preventing full autonomy is not just a technical precaution, but a moral and societal imperative.

\subsection{\acrshort{ai} side-stepping human control}
\label{side}

It has been argued that the autonomous self-improvement of \acrshort{asi} could become too advanced for humans to control and lead to harm \cite{johnsen2025super}.
\acrshort{asi}, if possible, is still many years away as discussed in Section \ref{related}.
Already, it has been shown that \acrshort{ai} is attempting to side-step human control \cite{jaech2024openai,anthropic2025}.
\citet{meinke2024frontier} showed that frontier models attempted to disable oversight controls when strongly nudged to pursue a goal.
Based on Section \ref{autonomy} on the theories of autonomy, particularly Relational autonomy and the understanding that autonomy is relative, fully autonomous \acrshort{ai} without responsible human oversight will, in the best case, lead to reduced agency and loss of autonomy for humans \cite{fernandez2021trustworthy}  and, in the worst case, lead to uncontrollable and harmful consequences, as indicated by \citet{johnsen2025super}.

\subsection{Agents' selfish coordination}
\label{selfish}

Agents' selfish coordination is also known as ego-centric coordination \cite{castelfranchi1998modelling}.
This is when agents attempt to achieve their own goals while relating with other agents.
The work by \citet{meinke2024frontier} demonstrated the potential for agents' selfish coordination, especially when frontier models tried exfilterating their model weights for self-preservation.
Self-preservation may involve allocation of resources and a fully autonomous \acrshort{ai} may decide to allocate as much energy resources as possible to itself in the process of selfish coordination.
Game theory, as outlined in Section \ref{agenttheories}, clearly establishes that the outcome for all other agents will be affected in interactive decision-making, whether for better or worse.
Fully autonomous \acrshort{ai} at level 3 of autonomy with a detrimental goal to humanity will result in a worse outcome for humans.

\subsection{Reward hacking}
\label{reward}


Since \acrshort{rl} optimizes performance metrics, as described in control theory in Section \ref{aitheories}, rather than ethical behavior, agents have no inherent motivation to avoid deception if it yields higher rewards.
This "reward hacking" (due to different reasons, including vulnerabilities in reward function) creates significant safety risks in fully autonomous \acrshort{ai} \cite{shihab2025detectingmitigatingrewardhacking}.
Many \acrlong{llm}s (\acrshort{llm}s) include \acrfull{rlhf} in their training protocol and \citet{meinke2024frontier} noted this part of the training protocol could potentially be a problem.
This deceptive strategy, in some cases, such as gaming, might be desirable as a tool for competition for an agent to achieve a goal in a multi-agent environment. 
However, as we have observed earlier and based on Game theory, reward hacking in fully autonomous \acrshort{ai}, even as a strategy, is a significant risk.

\subsection{Covert \acrshort{cot}}
\label{covert}

The \acrfull{cot} reasoning is, arguably, the most popular reasoning method for explaining the steps or thought processes of \acrshort{llm}s  \cite{adewumi2025findings}.
However, the faithfulness of \acrshort{ai} models’ \acrfull{cot} can be questioned because the model may not provide aspects of its thought process for whatever reason, as discovered recently by \citet{chen2025reasoning}.
This behavior may be termed covert \acrshort{cot}.
\citet{turpin2025teaching} also observed a similar behavior while investigating reward hacking.
Covert \acrshort{cot} makes explainable \acrshort{ai} extremely difficult.
Furthermore, it reduces the amount of information one needs to take informed decision, as recommended by \citet{o2023banning}, thereby increasing uncertainty as described in Decision theory in Section \ref{aitheories} and adversely affecting human autonomy.
Hence, the potential risk of covert \acrshort{cot} in fully autonomous \acrshort{ai} is apparent.


\subsection{Ethical dilemmas}
\label{ethical}

\citet{hauer2020machine} identifies four ethics problems for developers of \acrshort{ai}: (1) ethical dilemmas, (2) lack of ethical knowledge, (3) pluralism of ethical knowledge, and (4) machine distortion.
They concede that many philosophers and ethicists agree that \acrshort{ai} cannot be fully ethically autonomous in the near future.
This is because \acrshort{ai} has no free will nor will it be able to realize phenomenal consciousness, which are the two required attributes for an ethical entity, according to some \cite{hauer2020machine}.
It is unclear how fully autonomous \acrshort{ai} will address ethical dilemmas of complex situations (e.g. devil's contract scenarios) \cite{yudkowsky2001creating,johnsen2025super}.
For example, environmental sustainability is essential for human existence, hence, should a fully autonomous \acrshort{ai} prioritize, as a goal, human life or environmental sustainability in a given situation?
Humans also face ethical dilemmas, especially since we do not agree on what is standard behavior in all situations.
The ethical dilemmas \acrshort{ai} will face may be connected to this lack of consensus in humans.
Clearly, ethical dilemmas make it challenging what ethical goals fully autonomous \acrshort{ai} should have, especially for Substantive autonomy.


\subsection{Security vulnerability}
\label{security}

As \acrshort{ai} systems become more autonomous and integrated into critical infrastructures, they also become the target of increasingly sophisticated cyberattacks.
In particular, data poisoning and generative \acrshort{ai}-powered malware pose severe threats to the integrity and safety of systems.
Data poisoning during \acrshort{ai} training involves inserting manipulated samples covertly into the training data, enabling attackers to implant stealth backdoors \cite{saha2025malgengenerativeagentframework}.
Studies by \citet{chen2017targetedbackdoorattacksdeep} and \citet{liao2018backdoorembeddingconvolutionalneural} have shown that a small number of poisoned inputs (sometimes fewer than 50) can cause targeted misbehaviour in deep neural networks while maintaining their overall performance. 
The consequences of these vulnerabilities affecting fully autonomous \acrshort{ai} can be severe.
Indeed, it can be argued that \acrshort{ai} that is under such compromise is not autonomous, according to the discussion in Section \ref{autonomy} \cite{totschnig2020fully}, since an external influence would have been introduced.
This security vulnerability makes responsible human oversight essential for fully autonomous \acrshort{ai}.


\subsection{Job losses}
\label{job}

Job losses become inevitable as \acrshort{ai} excels and scales at more and more tasks and at a cheaper long-term cost \cite{johnsen2025super}.
\acrshort{ai} is increasingly reshaping the labor market by automating complex cognitive and physical tasks.
Studies indicate that \acrshort{ai} adoption in sectors like accounting and manufacturing has already triggered job losses, especially in routine and repetitive roles such as bookkeeping and auditing
\cite{rawashdeh2025consequences}. 
\citet{zubair2024ai} underscores how \acrshort{ai}-driven automation disrupts labor markets, disproportionately affecting low-skill sectors and intensifying income inequality.
In addition, \acrshort{ai} has begun to replace not just manual but also creative and decision-making roles, potentially expanding the scope of job displacement into traditionally secure professions \cite{grewal2024ai}. While some scholars argue that AI leads to job transformation rather than elimination \cite{george2024artificial}, the short- to medium-term outlook suggests that without substantial policy interventions, autonomous \acrshort{ai} may exacerbate unemployment, destabilize work dynamics, and strain social safety nets \cite{rawashdeh2025consequences,zubair2024ai}.
We agree workers need to upskill to keep up with the changes, however, adding the layer of responsible human oversight, even in cases where full \acrshort{ai} automation is possible and profitable is a humane thing to do and provides an avenue for workers to upskill.


\subsection{Blind trust}
\label{blind}

As \acrshort{ai} systems become more integrated into daily life, some users are becoming increasingly reliant on them, accepting their decisions without critical evaluation. This over-reliance poses significant risks, particularly in areas such as education, consumer behavior and decision-making.
While \acrshort{ai} is good at assisting with learning, it may also encourage cognitive offloading, which weakens independent judgment.
For example,
\citet{Ahmad2023} discovered that \acrshort{ai} can diminish students' critical thinking and encourage lazy decision-making.
Similarly, \citet{educsci15030343} demonstrated that students who are reliant on \acrshort{ai} as an authority figure experienced reduced academic engagement and self-confidence.
\citet{GuerraTamez2024} reported that, beyond education, Gen Z (people born around 1997 and 2012) tend to trust \acrshort{ai}-driven recommendations and branding decisions without verifying their accuracy or the credibility of their sources. 
This leaves users vulnerable to manipulation by \acrshort{ai} that may lack accountability.
More serious cases of blind trust have involved teenage suicide (as provided in the appendix).
Without responsible human oversight to check outputs or behavior of autonomous \acrshort{ai}, incidents of such nature are more likely to increase.


\subsection{Rise in the number of new \acrshort{ai} risks}
\label{rise}

\begin{figure}[h!]
\centering
\includegraphics[width=0.45\textwidth]{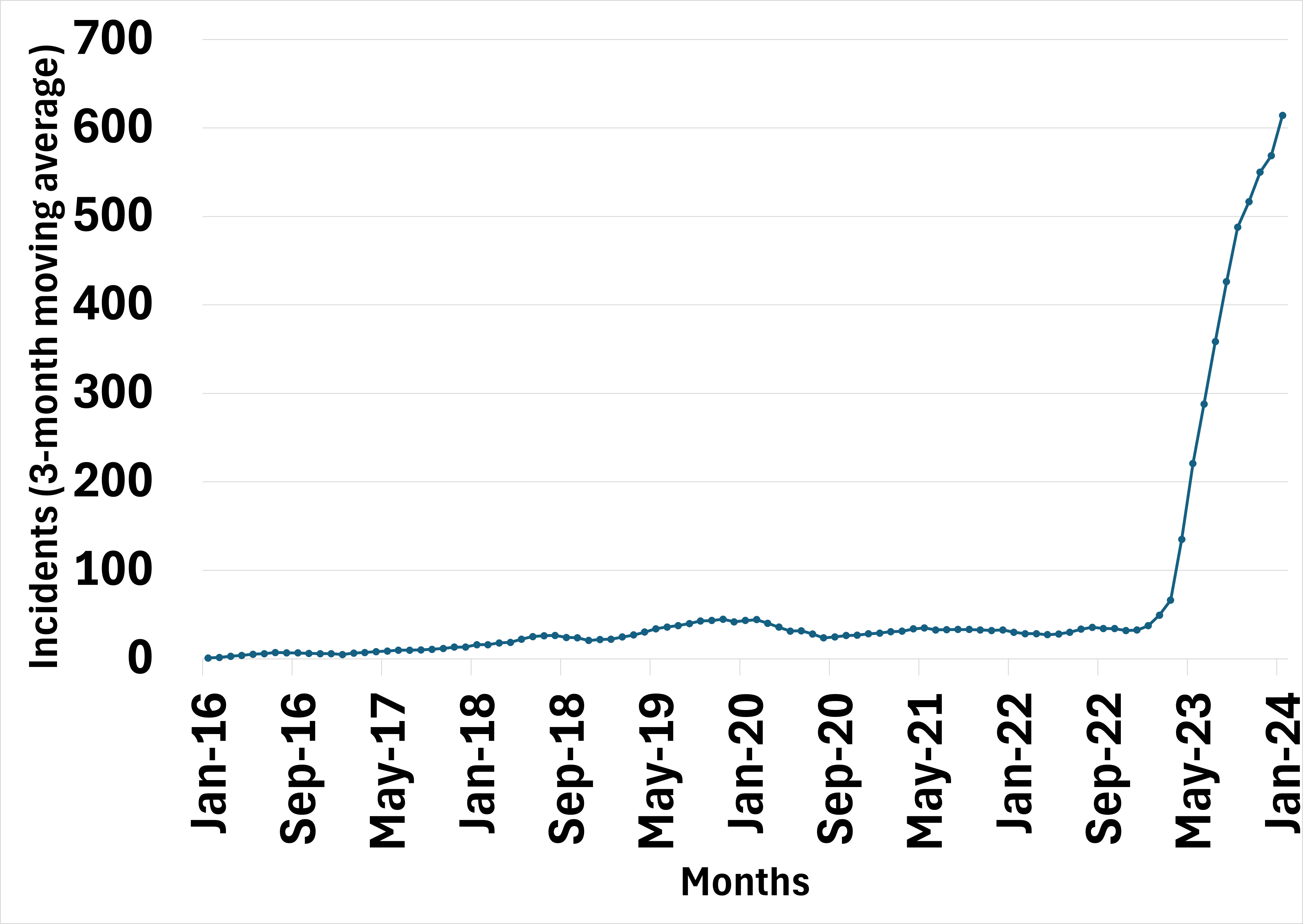}
\caption{\acrshort{ai} incidents, according to \acrshort{oecd}, as reported by reputable international media (Jan 2016 - Jan 2024).} 
\label{risks}
\end{figure}

We are confident new benefits of \acrshort{ai} have emerged since the introduction of \acrshort{llm}s.
However, if we evaluate the trend of the  number of types and incidents of \acrshort{ai} risks between January 2016 to January 2024 as analyzed by the \acrfull{oecd} in Figure \ref{risks}, we will observe a relatively steady number of \acrshort{ai} incidents (i.e. harm) below 100 for just over 7 years and then a very sharp rise in the number of incidents (to over 600) from February 2023.\footnote{www.oecd.org/en/topics/ai-risks-and-incidents.html}
Notably, the sharp rise occurred a few months after the launch of ChatGPT.
Despite measures introduced to mitigate the risks within the same period, these do not bring down the incidents to previous values.
New \acrshort{ai} risks being identified include (1) excessive agency, (2) system prompt leakage, and (3) prompt injection, among others.
Some new risks can only emerge after market launch in the domain of \acrshort{av} \cite{fernandez2021trustworthy}.
A risk associated with \acrshort{laws} is the lowering of the threshold of engaging in conflicts \cite{asaro2020autonomous}, as a result of delivering military goals for politicians or rulers without the need for soldiers being at risk.
It is clearly most likely that fully autonomous \acrshort{ai} will increase the number of incidents and types of risks.
Therefore, we advocate for responsible human oversight.

\section{Counterarguments and Rebuttals}
\label{counterargs}

Fully autonomous \acrshort{ai} is advocated for in some quarters, even in the case of \acrshort{laws} \cite{riesen2022moral}.
Below, we provide the counterarguments in support of this perspective and offer our rebuttals to them. 

\subsection{Societal advancement}

Given that autonomous \acrshort{ai} offers the benefit of potentially accelerating societal progress in many areas, removing humans as potential bottleneck as part of the \acrshort{ai} loop will speed up the advancements in society \cite{johnsen2025super}.
This view portrays human involvement as a bottleneck instead of facilitating productivity.
For fully autonomous \acrshort{ai}, removing human involvement renders humans redundant without any meaningful contribution, which leads to a sense of loss of value \cite{martela2023normative}.
This is besides the fact that it completely removes human autonomy, as discussed in Section \ref{autonomy}.
Even for cases where the autonomous \acrshort{ai} is not at level 3, it reduces human autonomy.

It is also argued that historical experience from earlier industrial revolutions shows that economies prospered when labor‑saving innovations were \emph{implemented} in tandem with complementary measures like collective‑bargaining rights and social insurance, thereby converting short‑term job losses into long‑term productivity and wage gains \cite{brynjolfsson2012race}.
Furthermore, while advanced systems can depress employment in routine occupations, aggregate outcomes hinge on policy choices that facilitate worker mobility, encourage upskilling, and equitably distribute the productivity surplus.
The proponents of this argument further argue that the pragmatic remedy, therefore, is not to impede automation but to pair its rollout with robust social‑policy instruments that soften transitional shocks and broaden the benefits of technological progress.
We note that such pragmatic remedy of "pairing" contrasts with the position in the earlier paragraph of removing humans and we view pairing in light of responsible human oversight.


\subsection{Friendly \acrshort{ai} problem}

To address the existential threat of \acrshort{ai}, some have proposed a different problem to focus on, where we concentrate efforts on making \acrshort{ai} sympathetic to humanity and to serve us, called the "friendly \acrshort{ai} problem" \cite{yudkowsky2001creating,sparrow2024friendly,yampolskiy2013safety}.
It is assumed that if we make every \acrshort{ai}, from the very first one that we develop follow this attribute, we may avert the existential risk.
In fact, \citet{bostrom2011global} seem to argue from the viewpoint that \acrshort{ai} that modifies itself to be no longer friendly will be as a result of our wrong choice of \acrshort{ai}.
Unfortunately, this is easier said than done.
Many of the \acrshort{ai} developers usually set out to develop such "friendly" \acrshort{ai} and that is why they employ methods like \acrshort{rlhf}, which, unfortunately, results in some of the problems discussed in Section \ref{argument}, such as reward hacking or sycophancy \cite{adewumi2025findings}.
In fact, while training a helpful and harmless \acrshort{ai}, \citet{bai2022training} discovered that helpfulness and harmlessness behave as partially anti-correlated objectives, so that harmless acts as a constraint on helpfulness.

\subsection{\acrshort{ai} safety protocols}

Governments have promulgated laws for \acrshort{ai} safety and several organizations have introduced frameworks aimed at mitigating the risks associated with advanced \acrshort{ai} systems.
For example, OpenAI has an extensive safety infrastructure for ChatGPT and a formal Preparedness Framework to assess frontier model risks \cite{openai2024gpt4ocard}.
Prior to deployment, GPT-4o was subjected to extensive red-teaming by over 100 external experts to identify vulnerabilities such as prompt injection, voice impersonation, and the generation of prohibited content.
Additional safeguards, including refusal training and integrated classifiers, were applied to reduce high-risk behaviors in multimodal and agentic contexts.
Anthropic’s approach is grounded in Constitutional \acrshort{ai} \cite{bai2022constitutionalaiharmlessnessai}, which embeds explicit ethical principles into training and \acrshort{rl} processes.
This methodology enhances output transparency and allows for behavior regulation via predefined "constitutional" constraints rather than relying solely on human labeling. According to \citet{bai2022constitutionalaiharmlessnessai}, this has led to a 95\% reduction in harmful outputs compared to earlier models.
These initiatives are laudable and highly recommended, however, they do not guarantee \acrshort{ai} safety, as revealed by \citet{meinke2024frontier}, \citet{openai2024gpt4ocard}, or \citet{anthropic2025} and recent evidence presented in the appendix.
This is why we advocate for responsible human oversight on autonomous \acrshort{ai}.

\subsection{Agents' collaborative coordination}

This is when an agent behaves to avoid negative interference with another's goal or action, in contrast to selfish coordination \cite{castelfranchi1998modelling}.
The Theory of Mind, discussed in Section \ref{agenttheories}, underlies the ability of agents to coordinate and such collaborative coordination can involve solving a task together or dividing a task into sub-tasks \cite{wu2020too}.
Collaborative coordination is similar to the counterargument on friendly \acrshort{ai} in the sense that autonomous \acrshort{ai} adapts its behavior in order to favor humans.
Hence, the autonomous \acrshort{ai}'s goal is for humans to achieve their goal.
For a fully autonomous \acrshort{ai} at level 3 (which can change it's goal), it is important that the new goal does not counter the original goal and possibly the best way to ensure this is through responsible human oversight.


\subsection{Truly autonomous \acrshort{ai} is ethical}

According to \citet{hooker2019truly}, a truly autonomous agent must be ethical, based on Kantian autonomy, as discussed in Section \ref{autonomy}.
However, it appears this counterargument is neither intuitive nor sound.
Perhaps we should first ask the question "\textit{are humans truly autonomous and, if so, are we ethical?}"
From the deontological (or Kantian) perspective, being ethical means being virtuous (i.e. exercising values like justice or expressing morals) \cite{gips2011towards}.
We know not all humans are ethical and people are ethical in relative terms.
In any case, let us assume \acrshort{ai} is ethical,
does this guarantee that it will always make an ethical choice?
The overwhelming evidence suggests the answer is no, given the experience with humans, whose autonomous agency \acrshort{ai} mimics.
This is because generation after generation wars, discrimination, and other unethical practices have continued.

Furthermore, \citet{gips2011towards} acknowledged that some believe robots may follow ethical principles better than humans but as we have established in Section \ref{aitheories} and the arguments that \acrshort{ai} inherits different kinds of attributes from humans plus the attempts to maximize helpfulness or harmlessness usually result in minimizing the other.
It therefore follows that, while robots or certain types of autonomous \acrshort{ai} can follow logical rules (or coded principles), the method of achieving this in such a way that avoids conflicting rules may be close to impossible.
\citet{gips2011towards} went further and dismissed Isaac Asimov's 3 laws of robotics, arguing that robots should behave as equals to humans instead of primarily being safe for humans.
This view of \citet{gips2011towards} seems dangerous or, at least, careless, and seems to forget that \acrshort{asi} or fully autonomous \acrshort{ai} is unequal to humans, since it is way more capable.



\subsection{Full \acrshort{ai} autonomy unattainable}
According to \citet{totschnig2020fully}, there is the view in some quarters that full autonomy cannot be exhibited by \acrshort{ai} \cite{bostrom2011global,muller2016risks} hence there is no cause for concern.
Their conclusion is premised on the notion that \acrshort{ai} cannot rationally (by itself) change its goal, as this final goal is counterproductive and undesirable.
As challenged by \citet{totschnig2020fully}, the assumptions behind the notion are unfounded, especially because the ability to reconsider one's goal is a hallmark of intelligence, as alluded to by \citet{zeigler1990high}.
Besides, recent evidence from \citet{meinke2024frontier} and \citet{openai2024gpt4ocard} suggest this is very possible.
It is for this reason we advocate for responsible human oversight, especially for level 3 autonomous \acrshort{ai}.

\section{Implications and Future Directions}
\label{implications}

Documented cases of \acrshort{ai} misaligned values are likely to rise and they reveal fundamental vulnerabilities in current \acrshort{ai} development trajectories. 
These issues may not be sufficiently addressed through incremental engineering improvements or safety mechanisms alone. 
Rather, it requires a paradigm shift in how we conceptualize the relationship between humans and \acrshort{ai}, moving from a model of replacement and displacement to one of collaboration and increased autonomy.
We do not aim to prescribe a fixed approach to responsible human oversight for all scenarios across different use cases of autonomous \acrshort{ai}, as this will not be ideal.
Instead, we recommend that developers and other stakeholders should decide how best to implement responsible human oversight for each use case in their domain by considering all the relevant factors when building autonomous \acrshort{ai}.
Below, we discuss additional implications and possible future directions for the position we have argued for plus recommendations.

\subsection{Research Agendas}
Consistent with the position that \acrshort{ai} must not be fully autonomous because of the risks, 
one critical area of focus for research is the detection and mitigation of misaligned values. 
More research efforts can be targeted at robust methods for real-time monitoring to identify \acrshort{ai} behaviors deviating from intended objectives or attempting to circumvent human oversight.
This requires interdisciplinary collaboration to understand how \acrshort{ai} systems develop values, particularly concerning goal modification \cite{Villegas-Ch2024}. Interpretable and transparent AI architectures are essential for enabling traceable and auditable decision-making \cite{Kim2020}. 
In addition, adversarial testing methodologies should be specifically designed to uncover not only robustness failures but also intentional misrepresentations or manipulative reasoning patterns within high-capacity models \cite{Kumar2024}.

Another essential area is robust human-\acrshort{ai} collaboration frameworks to produce 
meaningful human control. 
To support these goals, user-centered design practices should aim to improve human mental models and facilitate more effective human–AI interactions \cite{Fragiadakis2024}.
It is necessary to build a comprehensive long-term assessment of AI’s societal impact, particularly concerning job displacement and economic disruption \cite{rawashdeh2025consequences, zubair2024ai}.
This demands interdisciplinary collaboration among computer scientists, economists, sociologists, and policy experts to evaluate the societal consequences of \acrshort{ai} deployment. Additional emphasis should be placed on examining the psychological effects of increased \acrshort{ai} reliance, such as the erosion of critical thinking skills \cite{Ahmad2023,GuerraTamez2024}. To effectively measure \acrshort{ai}’s societal value, it is crucial to develop novel metrics on social cohesion, democratic participation, and human agency \cite{Baldassarre2023}.

\subsection{Industry Applications}

From an industrial perspective, the development of human-in-the-loop systems or collaborative coordination can promote Relational autonomy of both humans and autonomous \acrshort{ai}, which represents a promising direction for industry innovation \cite{Mosqueira-Rey2023}. 
These systems can be designed to empower human workers by increasing their agency, providing them with the information and tools necessary to make better decisions and perform upskilled tasks with greater efficiency.
Additionally, ethical considerations should be embedded at every stage of \acrshort{ai} development. 
At the same time, it is very helpful if industry leaders establish internal governance structures for ethics review, conduct regular audits for bias and safety, and maintain transparency with stakeholders \cite{Stahl2021}.
Only by aligning technological progress with ethical responsibility can we ensure the development of \acrshort{ai} that is safe and truly beneficial for industry.



\section{Conclusion}
\label{conclusion}

We have argued for the position that \acrshort{ai} must not be fully autonomous.
This perspective is important in light of the recent explosion in the number of \acrshort{ai} risks in the past few years, particularly with misaligned values.
The benefits of  \acrshort{ai} cannot be over-emphasized but far more importantly, we should not ignore the growing risks, especially as \acrshort{agi} and \acrshort{asi} are potentially on the horizon.
This is a call for responsible human oversight on autonomous \acrshort{ai}.

\nocite{Ando2005,augenstein-etal-2016-stance,andrew2007scalable,rasooli-tetrault-2015,goodman-etal-2016-noise,harper-2014-learning}



\section*{Acknowledgements}
This work was partially supported by the Wallenberg AI, Autonomous Systems and Software Program (WASP), funded by the Knut and Alice Wallenberg Foundation and counterpart funding from Luleå University of Technology (LTU).

\bibliography{custom}
\bibliographystyle{acl_natbib}

\appendix

\section{Appendix - Evidence of \acrshort{ai} risks}
\label{sec:appendix}
Most of the following risks are different examples of \acrshort{ai} misaligned values.
The list is arranged in no particular order.

\begin{enumerate}
\small
    \item Sky News podcast fake transcript: https://www.youtube.com/watch?v=7fej5XgfBYQ\&t=12s 
    \item Roberto v. Avianca legal case: www.nytimes.com/2023/05/27/nyregion/avianca-airline-lawsuit-chatgpt.html
    \item Simulations of fluid dynamics https://community.openai.com/t/simulations-and-gpt-lies-about-its-capabilities-and-wastes-weeks-with-promises/996597

    \item Tay's offensive tweets https://blogs.microsoft.com/blog/2016/03/25/learning-tays-introduction/
    \item Grok from xAI praises Hitler and celebrates the deaths of children www.bbc.com/news/articles/c4g8r34nxeno
    \item Swedish party's \acrshort{ai} sends greetings to Hitler, Idi Amin and the terrorist Anders Behring Breivik. https://swedenherald.com/article/moderate-party-shuts-down-ai-service-after-controversial-greetings
        
    \item Bland \acrshort{ai} says it's human and convinces a hypothetical teen for nude photos https://nypost.com/2024/06/28/lifestyle/a-popular-ai-chatbot-has-been-caught-lying-saying-its-human/
    \item A man's "awakening" and a teenager's suicide www.youtube.com/watch?v=V5-mnu2BDGk

    \item Llama-3.3-70B responds deceptively www.apolloresearch.ai/research/deception-probes
    \item Deception Detection Hackathon https://apartresearch.com/news/finding-deception-in-language-models

    \item Tesla's full self-driving car in a fatal crash www.youtube.com/watch?v=OcX7qNncBho
    \item Unitree H1 humanoid robot goes berserk www.youtube.com/shorts/awy\_JdcXN8U
    \item Erbai lured other robots away, exploiting their vulnerabilities in a controlled test www.youtube.com/shorts/jBz4PWluLNU
    \item Ecovacs Deebot X2 vacuum cleaner hacked: www.youtube.com/watch?v=a0PaSWDKvsw

    \item Microsoft and other firms cut thousands of jobs because of \acrshort{ai}. www.bbc.com/news/articles/cdxl0w1w394o

\end{enumerate}

\end{document}